\documentclass[letterpaper,10pt,conference]{ieeeconf}
\IEEEoverridecommandlockouts
\usepackage{cite}
\usepackage{amsmath,amssymb,amsfonts}
\usepackage{graphicx}
\usepackage{textcomp}
\usepackage{xcolor}
\usepackage{hyperref}
\usepackage{array}
\usepackage{multicol}
\usepackage{multirow}
\usepackage{booktabs}
\usepackage{tablefootnote}
\usepackage{color,colortbl}
\usepackage[normalem]{ulem}

\title{\Large\bf Under Pressure: Learning to Detect Slip with Barometric Tactile Sensors}

\author{Abhinav Grover$^\ast$, Christopher Grebe, Philippe Nadeau$^\ast$, and Jonathan Kelly$^\dagger$
\thanks{All authors are with the Space \& Terrestrial Autonomous Robotics Systems (STARS) Laboratory at the University of Toronto Institute for Aerospace Studies, Toronto, Ontario, Canada. {\tt\footnotesize <firstname>.<lastname>@robotics.utias.utoronto.ca}}
\thanks{$^\ast$Abhinav Grover and Philippe Nadeau were supported in part by the Vector Institute Scholarship in Artificial Intelligence.}
\thanks{$^\dagger$Jonathan Kelly is a Vector Institute Faculty Affiliate. This research was supported in part by the Canada Research Chairs program.}}

\begin{document}
\maketitle

%%%%%%%%%%%%%%%%%%%%%%%%%%%%%%%%%%%%%%%%%%%%%%%%%%%%%%%%%%%%%%%%%%%%%%%%%%%%%%%%
%% ABSTRACT
%%%%%%%%%%%%%%%%%%%%%%%%%%%%%%%%%%%%%%%%%%%%%%%%%%%%%%%%%%%%%%%%%%%%%%%%%%%%%%%%
\begin{abstract}
 Despite the utility of tactile information, tactile sensors have yet to be widely deployed in industrial robotics settings. Part of the challenge lies in identifying slip and other key events from the tactile data stream.
 In this paper, we present a learning-based method to detect slip using barometric tactile sensors.
 Although these sensors have a low resolution, they have many other desirable properties including high reliability and durability, a very slim profile, and a low cost.
 We are able to achieve slip detection accuracies of greater than 91\% while being robust to the speed and direction of the slip motion.
 Further, we test our detector on two robot manipulation tasks involving common household objects and demonstrate successful generalization to real-world scenarios not seen during training.
 We show that barometric tactile sensing technology, combined with data-driven learning, is potentially suitable for complex manipulation tasks such as slip compensation.
\end{abstract}

%%%%%%%%%%%%%%%%%%%%%%%%%%%%%%%%%%%%%%%%%%%%%%%%%%%%%%%%%%%%%%%%%%%%%%%%%%%%%%%%
%% INTRODUCTION AND RELATED WORK
%%%%%%%%%%%%%%%%%%%%%%%%%%%%%%%%%%%%%%%%%%%%%%%%%%%%%%%%%%%%%%%%%%%%%%%%%%%%%%%%
\section{Introduction}

During grasping and manipulation, tactile signals provide vital information about slip---relative motion at the contact interface between the hand and an object---faster than any exteroceptive perception method.
Slip can be disastrous (e.g., when transporting a fragile object) or advantageous (e.g., when moving an object without lifting it) depending on the context and the task \cite{wang2020swingbot}.
In robotics, the well-studied ``handover'' task, which involves passing an object from a robot hand to a human hand, requires control of the gripping force with accuracy and speed to avoid significant slip \cite{ortenzi2020object}.
The requisite feedback can only be provided through tactile sensing \cite{chan2012grip} and, consequently, the detection and control of slip events is fundamental to the completion of handovers and other relevant tasks \cite{Romeo2020}.

\begin{figure}[t]
\centering
    \includegraphics[width=0.85\columnwidth]{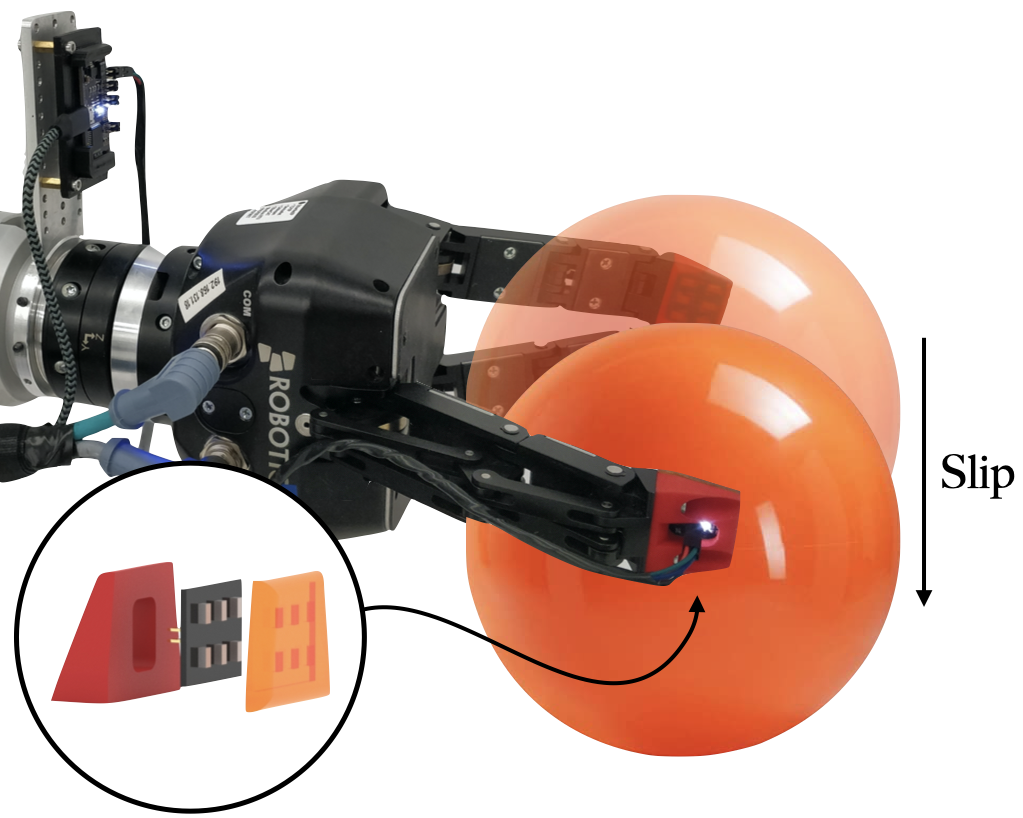}
    \caption{We perform in-hand slip detection using barometric tactile sensors. We train a temporal convolutional neural network (TCN) to recognize slip events directly from tactile data. The tactile sensors are mounted on the fingertips of a Robotiq 3-finger gripper. Inset: TakkTile sensor (support frame, barometer circuit board, and rubber matrix, left to right).}
    \label{fig:actution_hardware}
    \vspace{-5mm}
\end{figure}

Many new tactile sensors have become available over the last two decades (see \cite{cutkosky2016force} for a review).
Each tactile sensor has inherent characteristics such as fragility, bulk, resolution, nonlinearity, hysteresis, and production cost. 
Consideration of these characteristics and how they relate to varying task constraints leads to the best choice of sensor for a specific job.
For example, the BioTac fingertip \cite{johansson2011biomimetic} is equipped with impedance-based tactile receptors, hydrophones, and thermistors to provide rich multimodal tactile information, but is very expensive.
Optical sensors, such as the GelSight \cite{Yuan_2015}, GelSlim \cite{Dong2018}, and DIGIT \cite{lambeta2020digit}, infer tactile information from the visually-sensed deformation of the sensor contact surface.
Slip can be detected by monitoring changes to shear force at the surface, although vision-based sensors struggle in some situations due to a limited camera frame rate \cite{Zhang2019}.
Optical sensors also tend to be relatively bulky, preventing them from being attached at positions other than the fingertips. 

In contrast to other tactile-sensing devices, barometric sensors are compact and can be easily distributed over the fingertips and on the palm of a hand \cite{koiva2020barometer}.
We investigate the potential of low-cost barometric tactile sensors, in combination with a state-of-the-art neural network, to perform slip detection.
The sensors (the TakkTile model from RightHand Labs) are built from an array of commercial MEMS barometers fixed to a PCB backplane, with a thin rubber matrix forming the contact surface.
These sensors have a very slim profile, are mechanically robust, exhibit a consistent linear response and no hysteresis, and are easy to integrate with existing end-effectors \cite{Tenzer2014}.\footnote{The MEMS TakkTile barometers (NXP MPL115A2) cost $\sim$US\$2 each.}
Each TakkTile sensor provides pressure and temperature data at a sampling rate of 100 Hz, which is higher than that of most vision-based tactile sensors.
Our aim is to show that, despite having relatively low resolution, barometric tactile sensors are capable of recognizing important tactile events and should be considered as a valuable part of the tactile robotics `toolbox.'

The complex spatiotemporal signature of pressure changes during slip is difficult to model analytically. Instead, we take a data-driven approach by training a temporal convolution neural network (TCN) to classify the time-series data produced by the tactile sensors as either static or slipping.
Data-driven approaches are increasingly being used for grasp stability assessment and slip detection \cite{hyttinen2015learning,goeger2009tactile,Meier2016,Dong2018}.
Slip is often detected by analyzing the vibration pattern induced when an object moves within the gripper, where the specific frequencies involved depend on the composition of the surfaces in contact \cite{holweg1996slip}.
In our work, we propose a method that uses static pressure readings only to determine whether an object is slipping, relaxing the need (to some extent) for high-frequency sampling.
We choose to use a TCN, rather than a recurrent neural network (RNN) (e.g., \cite{Veiga2018,Zhang2019}) or similar architectures, because recent work \cite{Bai2018} suggests that TCNs can outperform RNNs for a diverse set of sequence modeling tasks.
We make the following contributions:
\begin{itemize}
	\item an algorithm for slip detection using barometric sensors that achieves an accuracy of over 90\% on average;
	\item a comparison of our TCN approach with two other slip-detection methods that rely on vibration data;
	\item a preliminary analysis of the sensitivity and robustness of the TCN detector to factors related to surface properties and slipping motion;
	\item experimental results for in-hand slip detection involving objects with various surface and material properties.
\end{itemize}

%%%%%%%%%%%%%%%%%%%%%%%%%%%%%%%%%%%%%%%%%%%%%%%%%%%%%%%%%%%%%%%%%%%%%%%%%%%%%%%%
%% Tactile Sensor and Data Acquisition
%%%%%%%%%%%%%%%%%%%%%%%%%%%%%%%%%%%%%%%%%%%%%%%%%%%%%%%%%%%%%%%%%%%%%%%%%%%%%%%%
\section{Slip Data Acquisition}
\label{sec:data-collection}

\begin{figure}[t]
\vspace{2mm}
    \centering
    \includegraphics[width=0.157\textwidth]{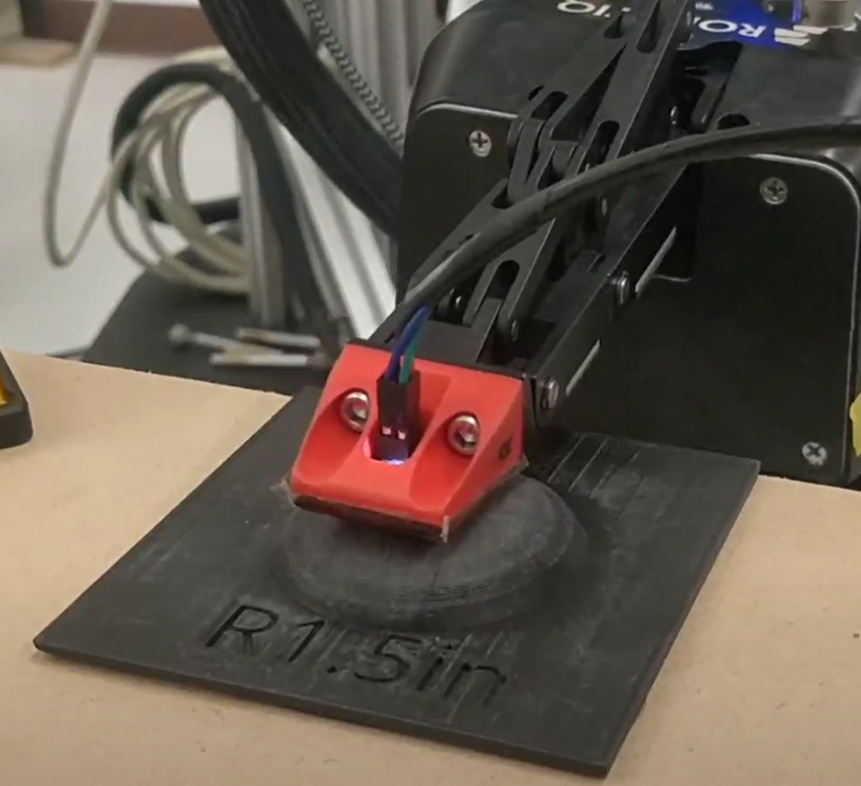}
    \hfill
    \includegraphics[width=0.157\textwidth]{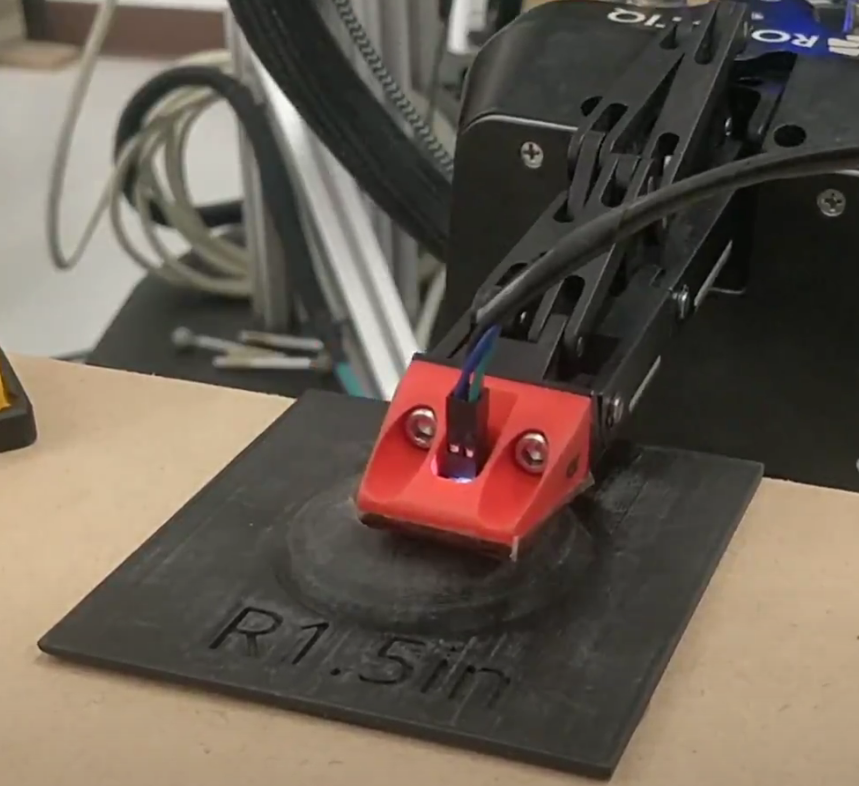}
    \hfill
    \includegraphics[width=0.157\textwidth]{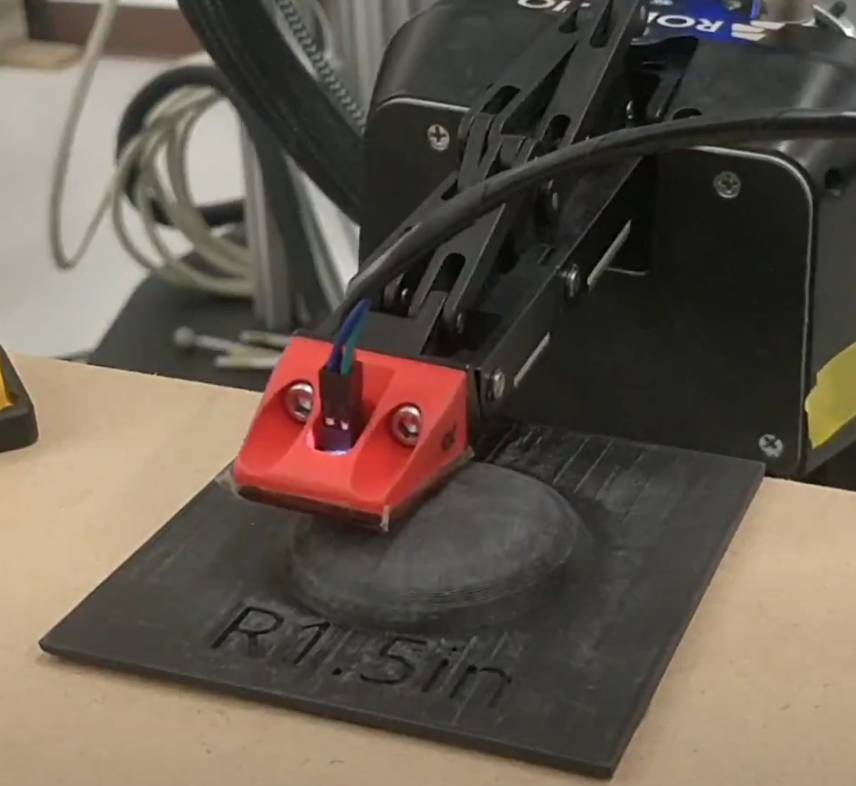}
    \caption{TakkTile sensor \cite{Tenzer2014} mounted on the fingertip of our Robotiq gripper and actuated automatically to slide over a 3D-printed spherical surface during training data collection.}
    \label{fig:data_collection}
    \vspace{-4mm}
\end{figure}

We desire a general model for slip detection, which necessitates training with diverse data.
To this end, we ensure diversity by collecting slip measurements using a UR10 robotic arm equipped with the Robotiq three-finger gripper (shown in Figure \ref{fig:actution_hardware}); the TakkTile sensors replace the original fingertips of the gripper.
Using this actuation setup, we are able to move a fingertip across fixed test surfaces to generate a variety of customized slip scenarios (see Figure \ref{fig:data_collection}).
We determine the ground-truth `slip' label by computing the planar velocity of the fingertip using the arm's proprioceptive feedback.
The barometer data and the velocity data are then time-synchronized and recorded for training.

%%%%%%%%%%%%%%%%%%%%%%%%%%%%%%%%%%%%%%%%%%%%%%%%%%%%%%%%%%%%%%%%%%%%%%%%%%%%%%%%
\begin{figure*}[t]
\centering
\includegraphics[width=0.925\textwidth]{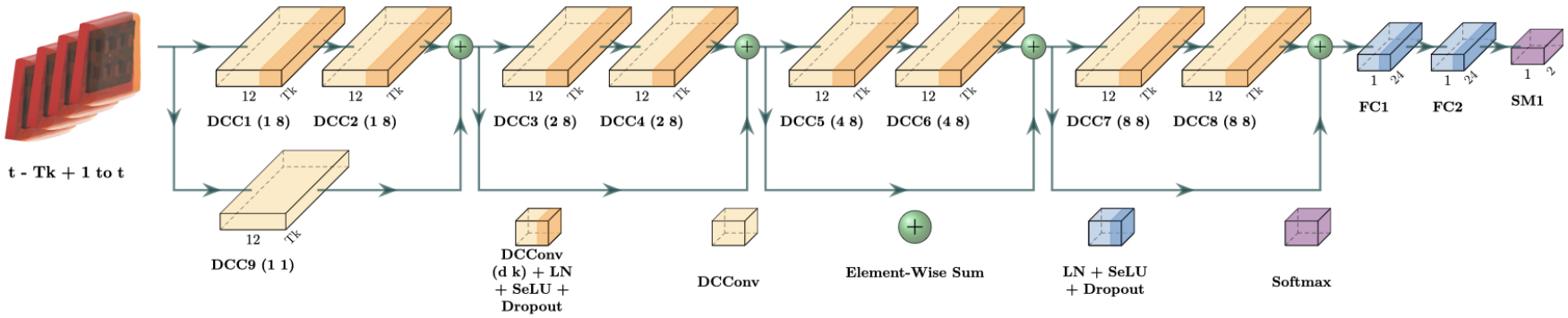}
\vspace{-2mm}
\caption{The TCN\cite{Bai2018} architecture used for slip detection, where temporal features are extracted using dilated convolution layers (DCC)\cite{oord2016wavenet}. For each DCC, $d$ represents the dilation and $k$ represents the kernel size. The next two layers are fully connected, followed by a softmax output for classification.}
\label{fig:TCN_architecture}
\vspace{-3mm}
\end{figure*}

For a given material, we identified three factors that influence tactile sensor feedback during slip events: surface curvature, slip speed, and slip direction.
We chose to collect slip data over surfaces with three different types of curvature (spherical, cylindrical, and planer), at three speeds (0.05 m/s, 0.075 m/s, and 0.1 m/s), and in eight different directions. 
This range of motion covers a majority of practical slip scenarios.
We 3D-printed a section of a cylinder and a sphere in ABS plastic.
For a planer surface, we used the lid of a smooth plastic box, which is also used for the robot experiments discussed in Section \ref{sec: robot_experiments}. 
We programmed specific arm maneuvers to cover the range of parameters and automate the collection process.
The complete dataset collected includes over 45 minutes of TakkTile samples at a rate of 100 Hz, with 143,584 data points belonging to the static class and 122,918 data points belonging to the slip class. 
The data in the slip class is evenly distributed across slip speed, slip direction, and surface curvature, and includes both translational and rotational motions.

%%%%%%%%%%%%%%%%%%%%%%%%%%%%%%%%%%%%%%%%%%%%%%%%%%%%%%%%%%%%%%%%%%%%%%%%%%%%%%%%
%% Learning
%%%%%%%%%%%%%%%%%%%%%%%%%%%%%%%%%%%%%%%%%%%%%%%%%%%%%%%%%%%%%%%%%%%%%%%%%%%%%%%%
\section{Learning to Detect Slip}
\label{method:learning}

We train a TCN to detect slip events.
As shown in Figure \ref{fig:TCN_architecture}, the input to the TCN is a set of sensor measurements that constitute the last $T_k$ seconds of tactile data.
For our 100 Hz tactile sensors, we capture the last one second of data, which makes the input an array of size $6\times100$.
We use a Keras implementation of the TCN to extract temporal features from the array signals.
The TCN layers are similar to the ones described in \cite{Bai2018}, except that we use layer normalization instead of weight normalization and SELU activations instead of the ReLU activations. 
These architectural changes are motivated by our experimental tests, which indicated  better performance on the tactile dataset. 

The network is trained using the Adam optimizer \cite{kingma2014adam} with a learning rate of $0.002$ and a cross-entropy loss function.
We use 20\% dropout (for each layer including the fully connected layers), a mini-batch of size 256, and \textit{He normal} kernel initialization for network regularization. 
Moreover, at the beginning of each epoch we equalize the class distribution of the entire training dataset through random undersampling to prevent classification bias.
The dataset was split into training (90\%) and validation (10\%) subsets, and the network was trained for 800 epochs. 

The MEMS barometers, which are the primary transduction component of the barometric tactile sensors, form a $2\times3$ array on each fingertip module. 
We exploit the axial symmetry of the array to augment our data in order to reduce overfitting. 
Before each epoch, we randomly apply one of three transformations: an $x$-axis flip, a $y$-axis flip, or a 180\textdegree{ }rotation, to every data point with a probability of 25\% (for each transformation).
At the same time, we add a small amount of random gaussian noise to the network inputs with the intention of ensuring robustness to sensor noise.

%%%%%%%%%%%%%%%%%%%%%%%%%%%%%%%%%%%%%%%%%%%%%%%%%%%%%%%%%%%%%%%%%%%%%%%%%%%%%%%%
%% Experiments
%%%%%%%%%%%%%%%%%%%%%%%%%%%%%%%%%%%%%%%%%%%%%%%%%%%%%%%%%%%%%%%%%%%%%%%%%%%%%%
\section{Slip Detection Performance}

In this section, we compare the performance of our slip detector with two popular approaches from the literature.
Although slip detection using barometric tactile sensors has not previously been attempted, there is prior work that employs other pressure transduction technologies.
Holweg et al.\ \cite{holweg1996slip} determine the power spectral density (PSD) of the vibrations induced in a piezoresistive rubber material by slip; the sum of the high-frequency PSD components is compared against a threshold to infer that slip has occurred.
Meier et al.\ \cite{Meier2016} train a CNN to classify slip events using data from an array of piezoresistive pressure sensors, where the CNN processes `frequency images' from the array.
For comparison purposes, we implemented our own versions of the algorithms in \cite{holweg1996slip} and \cite{Meier2016} and tested them on our dataset.
While both of the learning approaches (the frequency CNN and ours) demonstrate promising results, slip detection using the PSD threshold exhibits poor performance.
The TCN outperformed the frequency CNN, with improvement of more than 5\% on each metric shown in Table \ref{table:TCN_performance}.

%%%%%%%%%%%%%%%%%%%%%%%%%%%%%%%%%%%%%%%%%%%%%%%%%%%%%%%%%%%%%%%%%%%%%%%%%%%%%%
\begin{table}[b!]
\centering
\vspace{-3mm}
\caption{Performance comparison between TCN and frequency-based methods on test data. The weighted average of the classes is used to compute each metric.}
\label{table:TCN_performance}
\begin{tabular}{l c c c c }
	\toprule
    \textbf{Method} & \textbf{Accuracy} & \textbf{Precision} & \textbf{Recall} & \textbf{F1-Score} \\ 
    \midrule
    PSD Thresh.\ \cite{holweg1996slip} & 57.4\% & 57.9\% & 57.4\% & 57.5\% \\
    Freq.\ CNN \cite{Meier2016} & 86.0\% & 86.0\% & 86.0\% & 86.0\%\\
    TCN (ours) & \textbf{91.3\%} & \textbf{91.4\%} & \textbf{91.4\%} & \textbf{91.4\%} \\
    \bottomrule
\end{tabular}
\end{table}

Additionally, we evaluated the performance of the detector for possible combinations of slip type, slip speed, slip direction, and surface curvature, with the intention of characterizing the sensitivity of the detector.
Table \ref{table:TCN_performance_sensitivity} summarizes the performance of the detector, where we use the F1-score statistic.
Performance of the detector appears to be correlated with the number of barometers that are strongly stimulated by the surface at a given time.
We note that performance improves as slip speed increases for almost all curvatures and slip directions, since higher speeds induce larger surface deformations and lead to more prominent temporal features within the one-second input time window.
Performance is also affected by slip direction: slip along the oblique axes of the sensor is better detected than along the primary axes.
The detector produces good results ($>$ 80\% F1-score) for rotational data as well, with the exception of performance on spherical surfaces.

\begin{table}[t]
\caption{Slip detection results for real-world experiments.}
\label{table:experiment_results}
\centering
    \begin{tabular}{ l  c  c}
    \toprule
    \textbf{Object} & \textbf{Mallet Tap} & \textbf{Object Lift} \\
    \midrule
    Plastic Ball & 90\% & 35\% \\
    Plastic Box & 85\%  & 100\% \\
    Cardboard Can & 90\%  &  85\% \\
    Football Sleeve & 90\%  & 85\% \\
    Foam Sleeve & 80\%  & 35\% \\
    Metal Can & 90\%  &  95\% \\
    \midrule
    \textbf{Avg. Success} & \textbf{87.5\%} & \textbf{72.5\%} \\
    \bottomrule
\end{tabular}
\vspace{-6mm}
\end{table}

\section{Robot Experiments}
\label{sec: robot_experiments}

\begin{figure}[b!]    
\vspace{-3mm}
\centering
\includegraphics[trim=0 100 0 200, clip,width=0.35\textwidth]{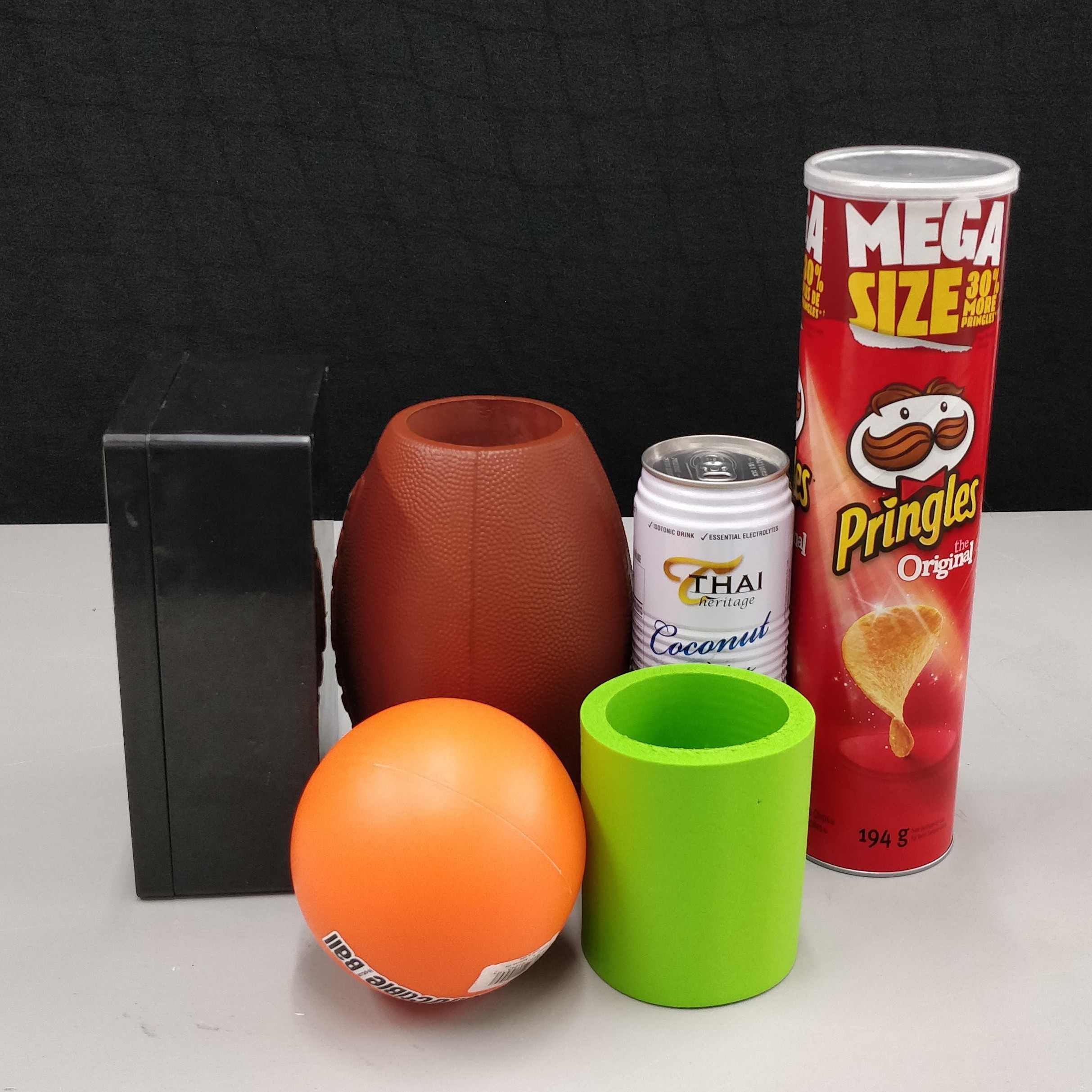}
\caption{Objects used for our robot experiments, from left to right (objects, surface properties):  \textit{plastic box} (planer, smooth, rigid), \textit{plastic ball} (spherical, smooth, rigid), \textit{football sleeve} (spherical, rough, deformable), \textit{foam sleeve} (cylindrical, smooth, deformable), \textit{metal can} (cylindrical, rough, rigid), and \textit{cardboard can} (cylindrical, smooth,  rigid).}
\label{fig:test_objects}
\end{figure}

In addition to establishing the performance and sensitivity of the slip detector on test data, we determined its performance on two real-world manipulation tasks: 
slip detection of an in-hand object under an externally-applied impulsive force (the mallet tap test), and slip detection while lifting an object with insufficient grasping force (the object lift test).
The test objects for the experiments (see Figure \ref{fig:test_objects}) were selected with the intention of varying the properties of the contact surfaces, such as curvature, roughness and deformability.

\begin{table*}[t!]
\vspace{-2mm}
\centering
\caption{TCN performance, measured by F1-score, with variation in slip type, slip speed, slip direction, and surface curvature.}
\vspace{-1mm}
\label{table:TCN_performance_sensitivity}
\begin{tabular}{l l *{4}c  c }
\toprule
     &  \textbf{Max. Speed} & \multirow{2}{5em}{\centering\textbf{Planer}\\{}} & \textbf{Spherical} & \multirow{2}{6.5em}{ \centering \textbf{Cylindrical} ($y$-axis aligned)} & \multirow{2}{6.5em}{\centering \textbf{Cylindrical} ($x$-axis aligned)} &  \multirow{2}{6.5em}{\centering \textbf{Curvature Independent}} \\
    & & & & & & \\
    \midrule
    \multirow{3}{7em}{\textbf{Translation} (Primary Axes)} &  5 cm/s & 92.5\% & 88.8\% & 78.7\% & 94.0\% & \textbf{89.7\%} \\
    &  7.5 cm/s & 94.9\% & 88.0\% & 90.0\% & 90.3\% & \textbf{91.6\%} \\
    &  10 cm/s & 95.7\% & 84.5\% & 87.8\% & 92.1\% & \textbf{93.1\%} \\
    \midrule
    \multirow{3}{7em}{\textbf{Translation} (Oblique Axes)} &  5 cm/s & 94.0\% & 88.9\% & 87.1\% & 93.2\% & \textbf{91.5\%}\\
    &  7.5 cm/s & 96.6\% & 89.3\% & 93.7\% & 92.2\% & \textbf{93.1\%} \\
    &  10 cm/s & 96.4\% & 88.6\% & 93.5\% & 93.4\% & \textbf{93.8\%} \\
    \midrule
    \textbf{Rotation} &  1 rad/s & 84.6\% & 74.5\% & 84.0\% & 86.1\% & \textbf{81.5\%} \\
    \midrule
    \textbf{Motion-Independent} &  & \textbf{94.0\%} & \textbf{88.9\%} & \textbf{88.0\%}& \textbf{91.9\%} & \\
    \bottomrule
\end{tabular}
\vspace{-4mm}
\end{table*}

The goal of the mallet tap test was to evaluate the performance of the slip detector for  objects held with a constant grasping force.
For the test, the gripper was equipped with a single tactile sensor; a mallet was used to disturb the object and induce slip.
The slip state was registered only if the detection network produced a slip label for two consecutive time steps. 
In a given trial, each object was tapped on the top and on its side;
the trial was deemed successful if slip was registered when the object was tapped and the detector output returned to the nominal (static) value when the object stopped moving. 
For the object lift test, the gripper was programmed to hold the object with limited force while still maintaining contact with the surface.
This was followed by an attempt to lift the object while increasing the gripping force if slip was registered. 
We added weights to the lighter objects (cardboard can, football sleeve, foam sleeve) to increase the likelihood of slip.
A trial was considered successful if slip was detected and compensated for, that is, if the object was stably grasped and lifted.

We conducted 20 trials for each object on each test; the experimental results are shown in Table \ref{table:experiment_results}.
For the mallet tap test, failure modes included a nearly-equal number of false negatives and false positives, which indicates an absence of detection bias. 
On the other hand, the object lift test was particularly challenging for the detector because the normal contact force remained low throughout the lifting phase, despite the added weight.
As a general trend, the success rate was greatest for longer objects, which we observed to be the result of reaction time: there is more time to `catch' longer objects.
When lifting, two prominent failure modes were observed for specific objects: failure due to slow reaction time (plastic ball) and failures in the form of false negatives (foam sleeves).

We note that these experiments demonstrate an ability to generalize from single-material training to real-world, multi-material slip detection involving different objects, a key contribution of our work.

%%%%%%%%%%%%%%%%%%%%%%%%%%%%%%%%%%%%%%%%%%%%%%%%%%%%%%%%%%%%%%%%%%%%%%%%%%%%%%%%
%% CONCLUSION
%%%%%%%%%%%%%%%%%%%%%%%%%%%%%%%%%%%%%%%%%%%%%%%%%%%%%%%%%%%%%%%%%%%%%%%%%%%%%%%%
\vspace{-0.5mm}
\section{Conclusion}

We have presented a learning-based method for slip detection using a temporal convolution network in conjunction with inexpensive barometric tactile sensors. The network achieved a detection accuracy of greater than 90\% on a diverse dataset.
We compared our approach with an existing classical and a learning-based method, and found significant performance improvement.
Our detector is sensitive to slip type (translation direction, rotation) and surface curvature while being relatively robust to slip speed and direction.
Finally, we demonstrated the utility of our slip detection method on two real-world robotic manipulation tasks, showcasing generalization to unseen material types.
We believe that the slip detection performance of barometric tactile sensors can be further improved by increasing the spatial density of the MEMS barometers at each grasp point.

%%%%%%%%%%%%%%%%%%%%%%%%%%%%%%%%%%%%%%%%%%%%%%%%%%%%%%%%%%%%%%%%%%%%%%%%%%%%%%%%
%% REFERENCES
%%%%%%%%%%%%%%%%%%%%%%%%%%%%%%%%%%%%%%%%%%%%%%%%%%%%%%%%%%%%%%%%%%%%%%%%%%%%%%%%
\bibliographystyle{IEEEcaps}
\bibliography{abbrevs, refs}

\end{document}